# Supply Restoration in Power Distribution Systems — A Case Study in Integrating Model-Based Diagnosis and Repair Planning

Sylvie Thiébaux    Marie-Odile Cordier
IRISA
Campus de Beaulieu
35042 Rennes Cedex, France
<Firstname>.<Lastname>@irisa.fr

Olivier Jehl    Jean-Paul Krivine
EDF-DER
1, Av. du Général De Gaulle
92140 Clamart, France
<Firstname>.<Lastname>@der.edfgdf.fr

## Abstract

Integrating diagnosis and repair is particularly crucial when gaining sufficient information to discriminate between several candidate diagnoses requires carrying out some repair actions. A typical case is supply restoration in a faulty power distribution system. This problem, which is a major concern for electricity distributors, features partial observability, and stochastic repair actions which are more elaborate than simple replacement of components. This paper analyses the difficulties in applying existing work on integrating model-based diagnosis and repair and on planning in partially observable stochastic domains to this real-world problem, and describes the pragmatic approach we have retained so far.

## 1 INTRODUCTION

The integration of model-based diagnosis and repair has mainly been studied in the context of applications for which it is *suboptimal* to completely identify the state of the system prior to repairing it [Friedrich and Nejdl, 1992; Sun and Weld, 1992]. The motivations are generally the following: observations are expensive and time-consuming, and prohibitive breakdown costs force us to take some repair actions urgently.

For some application domains, integrating diagnosis and repair is even more crucial because it is simply *impossible* to gain sufficient information to discriminate between several candidate diagnoses without carrying out some repair actions. This occurs when no sensor is available that enables us to observe the relevant data, or when sensors exist but may return erroneous information: since they modify the system's state, repair actions are the only means of acquiring additional information by confronting the available observations on the new state with expectations. Significant difficulties may arise in particular when repair plans for various candidates are incompatible, since we cannot be sure to choose an adequate repair plan before discrimination, but also cannot discriminate further before part of the plan has been executed.

Restoring supply in a faulty power distribution system, which is a major concern for electricity distributors, is such an application. While the cost of observations is not an issue here, it features various types of uncertainties such as missing information, erroneous information, and stochastic actions which are more elaborate than simple replacement of components. Furthermore, different candidate diagnoses require subsequently different repair plans. We found that existing work on integrating model-based diagnosis and repair [Friedrich and Nejdl, 1992; Sun and Weld, 1992], as well as work on planning in partially observable stochastic domains [Cassandra et al., 1994; Draper et al., 1994], are unable to solve the problems raised by this application because the formalisms and methods used are not powerful enough or computationally too expensive.

The purpose of this paper is twofold. Firstly, and after introducing the problem of supply restoration in power distribution systems operated by the French electricity utility Electricité de France (EDF) (Section 2), we find it useful to explain the difficulties we encountered in applying existing research on planning in stochastic domains (Section 3) and on integrating model-based diagnosis and repair (Section 4). Secondly, we describe the pragmatic approach we have retained so far (Section 5). Since this latter sacrifices generality and solution optimality for the sake of efficiency, we hope that our conclusions (Section 6) will motivate further research in the two mentioned communities.

## 2 THE CASE

### 2.1 EDF DISTRIBUTION SYSTEMS

A power distribution system, as in Figure 1, can be viewed as a network of electric lines connected via switching devices (SDs) and fed via remote controlled circuit-breakers (CBs). SDs and CBs have two possible positions: either open or closed. A CB supplies power iff it is closed, and an SD stops the power propagation iff it is open. The positions of the devices are set so that the paths taken by the power of each CB form a tree called feeder. The root of a feeder is a CB, its leaves are open SDs, and each line belongs to a single feeder. Consumers may be located on any line, and are then only supplied when this line is fed. In the figure, CBs are represented by large squares, remotely controllable SDs (e.g. RSD 12) by middle-sized squares, and manually controllable SDs (e.g. MSD 10) by small squares. White devices (e.g. RSD 63) are open, and the others are closed. Adjacent feeders are distinguished using different gray levels, and the indexes of the devices located on the same feeder have the same first digit.

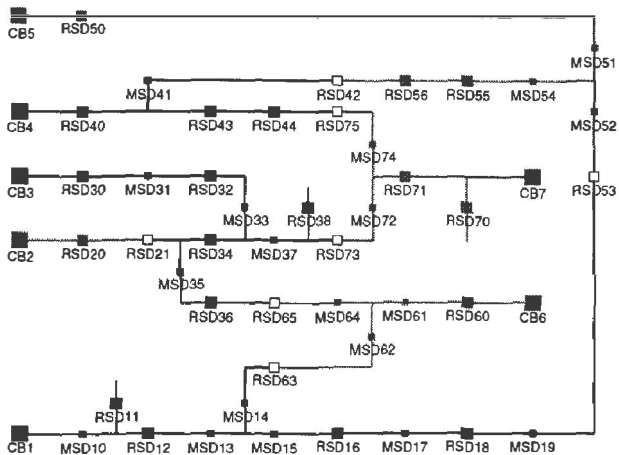

Figure 1: Power Distribution System (for semi-rural areas – representative data supplied by EDF)

In case of a permanent fault (short circuit) on a line, the CB feeding this line opens in order to protect the rest of its feeder from damaging overloads. A few minutes are then available to locate the faulty area,[1] to isolate it by opening the remote controlled devices surrounding it, and to restore the supply to the non-faulty areas by opening and closing remote controlled devices so as to direct the power towards those areas. This is a diagnosis and repair problem. The repair phase amounts to building a restoration plan consisting of switching (opening/closing) operations.

At present, this task whose steps are detailed below is carried out by an operator on the basis of his expertise and of pre-established restoration plans. Both are specific to a particular "normal" configuration of a particular network (the configuration in which this network is normally operated). EDF studies the automatization of this task, in order to improve its speed, the quality of the restoration plans, and the treatment of large extent incidents (e.g., multiple faults) which force moving away from the normal configuration.

---

[1] By area, we denote a set of lines surrounded by remote controlled devices. These areas constitute the finest possible granularity of remote diagnosis.

## 2.2 FAULT LOCALIZATION

Fault detectors (FDs) situated on the RSDs are the basis on which the fault localization phase is carried out. These permanently indicate whether or not they have "seen a fault pass them", i.e., whether or not a fault is downstream on the feeder. Then ideally, the fault is in the area between a sequence of RSDs whose FDs indicate that it is downstream and a sequence of RSDs whose FDs indicate that it is not. For instance, suppose that the fault is located on the area between RSDs 16 and 18, then CB 1 opens because it feeds this area, and the FDs of RSDs 12 and 16 should indicate that the fault is downstream while those of RSDs 18 and 11 should indicate that it is not.

Unfortunately, FDs can be broken (i.e., they do not return any information), or even lie (i.e., they return erroneous information). It follows that several hypotheses of faulty areas exist, each of which corresponds to an hypothesis concerning the behavior mode of the FDs. For instance, suppose that CB 1 opens and that the FD of RSD 16 indicates a fault is downstream while those of RSDs 12, 18, and 11 do not indicate a fault. Even assuming a single fault, there are 5 hypotheses: either the fault is upstream of RSDs 12 and 11, in which case 16 lies, or it is downstream of 11, and 16 and 11 lie, etc... There exist preferences between these hypotheses (the probability of multiple faults is much smaller than that of an FD lying, and this latter is higher when the FD indicates a fault downstream than when it does not because FDs do not detect all types of faults), but only the repair phase may enable us to discriminate, especially when this one goes wrong.

## 2.3 POWER RESTORATION

The suspected area is isolated by opening the remote controlled devices surrounding it. Then, the non-faulty areas are resupplied by operating remote controlled devices so as to direct the power towards those areas. For instance, if we suspect the area between RSDs 16 and 18, we may open 16 and 18 to isolate it, then reclose CB 1 to resupply upstream lines, and last close RSD 53 to resupply downstream lines via CB 5.

The following constraint determines which restoration plans are admissible: CBs and lines can only support a certain maximal power. This might prevent directing the power through certain paths and resupplying all the non-faulty areas. Ideally, restoration should optimize certain parameters under this constraint, such as resupplying as many consumers as possible, (giving priority to critical consumers like hospitals), minimizing the number of switching operations so as to stay close to the normal configuration, and balancing power margins of CBs in anticipation of the next load peak.

Even if the fault localization is correct, supply restoration is rendered difficult by the unreliability of the actuators (AC) of the remote controlled devices. An AC can be broken (it fails in executing the switching op-eration and sends a negative notification), or it can lie (it fails in executing the operation but sends a positive notification). In many cases of positive notification, it is still possible to know whether the operation has been executed or not by consulting the position detector (PD) of the device, which indicates whether this latter is open or closed. However, PDs can be broken (they do not return any information), in which case uncertainty remains. When a CB opens during the restoration process, it is then difficult to know whether this is due to a wrong fault localization, to a second fault which could not be detected, or to the failure of a switching operation meant to isolate the fault.

Note that it is reasonable to assume that the behavior mode of an FD does not change during the restoration process. However, that of a PD can change at anytime, and that of an AC can evolve at anytime from correct to liar or broken with given probabilities, the two abnormal modes being permanent. Since we are only interested in the modes of the ACs of the devices on which we perform switching operations, this amounts to considering these latter as stochastic actions that may change the behavior mode of the AC in addition to changing (or not) the position of the device.

### 2.4 MAIN FEATURES OF THE CASE

The problem of supply restoration in power distribution systems features numerous sources of uncertainty[2] due to partial observability, i.e. both incomplete and erroneous information about the current state of the network throughout the restoration process, and to stochastic actions. While the available sensing information is free, acquiring the missing information and identifying the erroneous one require executing some of the repair actions and confronting the result with expectations. Furthermore, since different fault locations and different behavior modes of the ACs require subsequently different restoration plans, the currently executed plan will have to be revised when it turns out to be inappropriate.

Additional difficulties are the following. Firstly, it is impossible to formulate a precise repair goal to be achieved, since we would have to know in advance which lines can be resupplied. Instead, we want to optimize plan utility, taking into account the parameters mentioned above. Secondly, plan evaluation should ideally take into account the risks in case of failure, the breakdown costs being potentially high. Thirdly, actions are far more complex than simple replacement of components, and have numerous ramifications which depend on the execution context (for instance, closing an SD may result in several lines becoming fed and even to a CB opening if a newly fed line is faulty).

---

[2]Power *transmission* systems have already been studied in the literature [Mondon et al., 1991; Friedrich and Nejdl, 1992; Beschta et al., 1993]. Their features are quite different, in that only a very few sources of uncertainty need to be considered.

Finally, the state space is huge. For instance, the network in Figure 1 has about $2.10^{67}$ states. The space of admissible restoration plans is huge as well, which makes the selection of a good plan without generating them all problematic.

The properties of this application make it non-trivial to design a model-based diagnosis/repair system. In a first step, EDF has built AUSTRAL, a prototype integrating a special-purpose model-based reasoner (for localization of the fault based on initial discrepancies, update of the state of the network upon occurrence of an action, and verification of the admissibility of plans), and an expert system (for plan selection/revision and further hypothetic reasoning) [Bredillet et al., 1994]. The hypothetic reasoning performed by the expert system is not systematic, since a failure of the current plan only leads to a revision of the fault localization hypothesis or to an abortion of the restoration. Also, the AUSTRAL prototype is limited to a single fault, and plan evaluation does not account for the consequences of possible failures. A second step was then to investigate recent developments in planning and model-based diagnosis which could enable us to overcome some of these limits.

## 3 PLANNING TECHNOLOGY

As noted in [Sun and Weld, 1992, p. 70], a first approach to integrating diagnosis and repair is to rely solely on the planning technology: a general-purpose planner coping with the types of uncertainties present in the application is used for both the diagnosis and repair tasks. It is obvious that, in the planning terminology, we are faced with the problem of acting optimally in a *partially observable stochastic domain*. Two types of works dealing with such domains have emerged in the planning literature, none of which turns out to be adequate for our problem, as we now explain.

The first one starts with the traditional techniques from operation research for solving partially observable Markov decision processes (POMDP), and focuses on improving and adapting them to the AI perspective [Cassandra et al., 1994]. This work is attractive in that the POMDP model is general enough to encode our problem, and even though the domain representation issues have not yet been addressed within such approaches, we can imagine formalisms that could make this encoding concise [Thiébaux et al., 1993]. Unfortunately, the currently available algorithms for solving POMDPs potentially explore the whole *belief* state space, which clearly makes our application out of their scope from the point of view of time-complexity.

The second type of works starts with the traditional representations and algorithms from AI-planning and extends them to account for stochastic and information-gathering actions [Draper et al., 1994]. The main advantage of this approach over the previous one is that the belief state space is only very implicitly explored. Nevertheless, we see two major difficulties in applying such a framework to our problem. At present, it addresses only a subclass of POMDPs for which plan utility is measured as the probability of satisfaction of a very precise goal, which must exceed a certain threshold. As explained above, we cannot express such a goal, and our needs in terms of plan utility are quite elaborate. More importantly, the framework does not yet account for domain constraints, and hence the descriptions of the actions must enumerate all ramifications in all contexts. This is impractical for our application, since this would make the size of the descriptions of switching operations exponential in the number of devices. It is not yet clear how the *algorithms* in [Draper et al., 1994] could be extended in any of the two directions and still keep their advantages over those in [Cassandra et al., 1994].

In order to experiment with the idea of grounding a supply restoration system on a general purpose planner, we decided, even though this solution only partially accounts for the risks in selecting a plan, to rely on PASCALE, a planner for *fully* observable stochastic domains developed by us [Thiébaux et al., 1993]. Roughly, PASCALE generates partial stationary policies as in [Dean et al., 1993], and is based on a more powerful action formalism than the BURIDAN representation [Kushmeric et al., 1994]. Notably, it allows the inference of ramifications via domain constraints.

The main positive results in the use of PASCALE are that (1) domain constraints keep the specification of the switching operations very concise [Thiébaux, 1995, chap. 6], and (2) compared to the AUSTRAL prototype, hypotheses are handled in a systematic way, multiple faults are coped with, and plan evaluation accounts for some of the risk factors. The major negative result is that PASCALE can only cope with toy networks. Two factors largely contribute to this. Firstly, PASCALE's treatment of domain constraints is far too powerful for the needs of the application. Second, PASCALE does not provide ways of specifying heuristics exploiting key properties of the application (the locality of the restoration, the tree structure of the network, the independence of most of the switching operations) which could considerably reduce the search space.

At present, we have no conjecture as to whether remedying to those problems would be sufficient for PASCALE to cope with real-size networks. In particular, domain constraints *are* needed for this application and are expensive to handle anyway. However, we strongly believe that no other existing general-purpose planner powerful enough to encode the fully observable aspects of the application [Dean et al., 1993] or even less [Kushmeric et al., 1994] would have performed significantly better than PASCALE. Since planners for fully observable domains only provides us with an upper bound on the size of the problems accessible to planners for partially observable domains, we conclude that the current planning technology is too expensive or not powerful enough for our application, and maybe both.

## 4 DIAGNOSIS TECHNOLOGY

A second approach to integrating diagnosis and repair which is clearly advocated by the model-based diagnosis community relies on a two-level architecture. At the top level, a diagnostic reasoner maintains a probability distribution on the candidate set, and chooses, at each step, whether it is preferable to discriminate between several candidates or to undertake some repair activities, according to breakdown, observation, and repair costs. At the lower level, a classical planner is responsible for returning an action sequence achieving a given repair goal for a given candidate. Upon executing an action, resp. obtaining new observations, the diagnostic reasoner updates, resp. revises, the candidates set. It turns out that the two works based on this approach [Friedrich and Nejdl, 1992; Sun and Weld, 1992] are not powerful enough for our application, as we now explain.

[Friedrich and Nejdl, 1992] essentially describes algorithms supporting the interleaving by the diagnostic reasoner of pre-established observations procedures and repair plans. At each step, these algorithms partition candidates into clusters, in such a way that *not* discriminating between candidates in a cluster and executing a repair plan resulting from somehow merging the individual plans for these candidates be preferable to discriminating and executing the individual plan for the remaining candidate. When the process stops, the system's state has been completely identified. This approach has been designed to address a significantly easier problem than those raised by our application: given that every relevant observation can be made reliably when needed, that repair actions are reliable, and that the repair plans for the various candidates are compatible enough to be merged (this only holds for basic repair actions such as component replacements), find a good interleaving of observations and repair actions, wrt. breakdown, observations, and repair costs.

In essence, the IRS system [Sun and Weld, 1992] has been designed to solve the same problem, but its features make it closer to our application's needs. At each step, the diagnostic reasoner chooses the best diagnostic goal among probe goals (finding out some information) and repair goals (reestablishing a desired functionality). This is done by projecting each possible choice on the candidates in the probability distribution and evaluating its consequences over a given horizon. The first few actions produced by the planner for achieving the best diagnostic goal are executed, and the whole process restarts until the reliability of the system exceeds a given threshold. IRS is based on the UWL language in which diagnostic goals, actions and states are described and used as input by the planner.

For our problem, the main advantage of IRS over the approach in [Friedrich and Nejdl, 1992] is that UWL does not make any distinction between observations and repair actions. Similarly, the diagnostic reasoner treats probe and repair goals uniformly. Hence, there is no requirement that some observations be taken before acting. The most important limitation of IRS is the absence of domain constrains in UWL and the associated planner. Though domain constraints are expensive to handle, giving them up in the planner implies either committing to very basic repair actions (IRS is explicitly limited to component replacements), or, as mentioned in Section 3, adopting very intricate and prohibitive action descriptions.

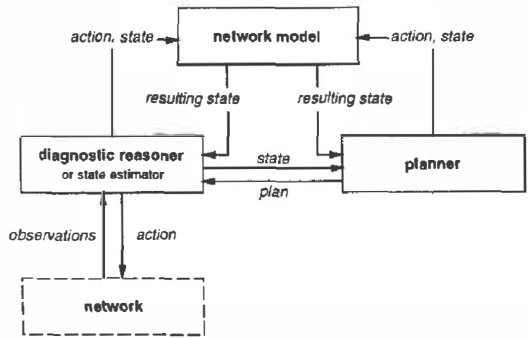

Figure 2: Architecture of the Prototype

Thus, the two-level architecture separating diagnosis and planning does not eliminate the need of dealing with domain constraints imposed by real-world applications. Its real advantage over the sole use of a planner is rather that this separation and the projection over a given horizon provides us with a flexible way of interleaving computation (of an approximately optimal plan) and execution, revising the plan as soon as observations puts its adequacy into question.

## 5   OUR CURRENT PROTOTYPE

Despite our problems with applying existing work, we had to come up with a prototype that resolved some of the issues left open in AUSTRAL and that could still be used on real-size networks. Our choices were dictated by the need of evaluating, in the short term, the feasibility of coping with real-size networks at all: the requirement that power eventually had to be reestablished was given priority over any other concerns. To avoid handling domain constraints and to fully exploit the properties the problem, we decided to sacrifice generality and to build an entirely domain-dependent prototype. Since coping with partial observability at planning time is expensive, we decided, at the cost of optimality, to plan as if the domain was fully observable and to revise the executed plan when needed. Therefore, and to facilitate a progressive evolution towards a better treatment of partial observability in the middle term, we chose to rely on the two-level architecture.

This architecture is shown in Figure 2. A domain-specific model similar to that found in AUSTRAL accounts for both the logical and quantitative aspects of power distribution systems. This model can be viewed as a network simulator, which returns the state resulting from performing a given switching operation on a given state of the network.[3] It can also be questioned to return the available sensing information and indicate whether an admissibility constraint (maximal power on a line or a CB) is violated. The diagnostic reasoner mainly acts as a domain-specific state estimator, which accounts for the history of actions and observations. It computes an initial probability distribution on states, asks the planner for a restoration plan for the most probable state, starts the execution of this plan — using the network model for updating the probability distribution upon execution of a switching operation, and revising it according to the new sensing information — and asks the planner for a new plan whenever the current one is inappropriate for the most probable state of the new distribution. The domain-specific planner uses the network model to compute an admissible restoration plan for the state hypothesis given by the diagnostic reasoner.

We now detail the principles underlying the diagnostic reasoner and the planner. We then present an example session with our prototype and briefly indicate how we will extend this latter.

### 5.1   THE DIAGNOSTIC REASONER

The three main tasks of the diagnostic reasoner are (1) to compute an initial probability distribution, (2) to update and revise this latter upon executing an action and obtaining new observations, and (3) to decide when to ask the planner for a new plan to be executed.

The first task takes as input the initial configuration of the network before the incident, the set of feeders which have been cut off, the information returned by the FDs of the RSDs on those feeders, and the maximal number $k$ of faults per feeder to be considered. It computes the behavior modes of these FDs that explain this information for each combination of at most $k$ faulty area hypotheses on each of those feeders.[4] This is done as follows. The network model is used to compute the states resulting from introducing each combination of faults on the initial configuration of the network, and to gather the sensing information that the FDs should produce for this combination if they were all correct. The comparison of this sensing information with the actual one determines the behav-ior modes of the FDs in the state associated with the combination: each FD that does not actually return any information is broken, each FD that actually says that the fault is downstream while it should say that it is not (or vice versa) is lying, and each other FD is correct. Since the probability of the mode of each FD only depends on the information it returns, the initial probability of a state is then simply the normalized product of the probabilities of the behavior modes of its FDs given the information they return.[5]

---

[3]We consider complete states, including the positions of the devices, the power on the lines, the faulty areas, and the behavior modes of the various sensors and actuators. Note that the network model considers that switching operations are deterministic, and stochasticity is handled by the diagnostic reasoner (see Subsection 5.1).

[4]Since the probability of multiple faults is much less than that of FDs lying, and since generating all combinations of an arbitrary number of faults per feeder is computationally too expensive, we start by considering a single fault per feeder. If it turns out later in the restoration process (see below) that there must be at least two faults on at least one feeder, then all combinations of two faulty areas on one of the feeders and at most two on the others are examined, and so on, for an arbitrary number $k$ of faults.

We now turn to task 2. The most probable state of the distribution is given to the planner which returns a plan for it. After the execution of a switching operation in this plan, the new distribution is computed as follows. We first *update* the old distribution: for each state and each possible change of behavior mode of the AC concerned by the switching operation, we use the network model to compute the state resulting from this operation and this change of behavior mode, and we transfer the probability mass of the formers to the latter. Then, we *revise* the updated distribution using Bayesian conditioning on the actual sensing information: this amounts to pruning the states for which the sensing information expected by the network model is inconsistent with the actual one, and normalizing the remaining probabilities. In the case where all states are inconsistent with the observations and have to be pruned, showing that the number $k$ of faults per feeder considered was to small, the diagnostic reasoner increments $k$ and restarts task 1.

We finally examine task 3. After each execution of a switching operation, the current plan is still considered as adequate if the most probable state in the new distribution is that which was expected by the planner. Otherwise, the planner is asked for a new plan starting from the now most probable state, and task 2 restarts with the first action in this new plan. The diagnosis/repair process ends when there is no remaining action to be executed in the current plan.

## 5.2 THE PLANNER

Plans returned by the planner consist of two sets of actions on remote controlled devices: opening operations (e.g., in order to isolate the faults), and closing operations (e.g., in order to restore the power to non-faulty lines). All opening operations must be performed before any closing one, but there is no other constraint. The number of such plans being exponential in the number of remote controlled devices, we restrict the search space to so-called level-1 plans: plans that only extend existing feeders, i.e., do not discharge any CB of part of its load after the incident. The space of admissible level-1 plans for the state hypothesis provided

---

[5]ACs are assumed correct in each state of the *initial* distribution, but not later on. PDs are always assumed correct in all distributions. Indeed, finding out their actual mode (correct or broken) from the observations is trivial and does not affect our decisions.

```
function plan(N, Cutoff) =
/* N : hypothesis about the state of the network */
/* Cutoff : set of feeders that have been cut off */
X ← ∅  /* devices at which to extend a feeder */
for all f in Cutoff
    for all extremity d of f
        /* common leaves with the bordering feeders */
        if d is still fed on another feeder f' then
            add (d, f') to X
        /* CBs of the cut of feeders */
        if d is a CB then
            add (d, f) to X
/* generate all admissible level-1 plans */
explore(N, ∅, ∅, X)
return the best plan found after evaluating them all

procedure explore(N, Open, Closed, X) =
/* Open/Closed: choices made in the current plan */
/* choice of the position of a device in X */
if X is not empty then
    let (d, f) be an element of X
        case
            /* position already chosen: keep it */
            d ∈ Closed: Choices ← {closed} break
            d ∈ Open: Choices ← {open} break
            /* inoperable device: choose current position */
            d is a MSD, or its AC is incorrect in N:
                Choices ← {position(d, N)} break
            others: Choices ← {open, closed}
        /* explore the choice of stopping the extension */
        if open ∈ Choices then
            explore(N, Open ∪ {d}, Closed, X \ {(d, f)})
        if closed ∈ Choices then
            /* explore the choice of going on the extension */
            if closing d in N does not lead to violate admissibility, a faulty line to be fed, or a line to be fed via multiple CBs then
                explore(N, Open, Closed ∪ {d}, (X \ {d, f}) ∪
                    {(d', f) | d' ∈ children(d, f)})
    return
/* when X is empty, an admissible plan is found */
/* remove the choices that are already satisfied in N*/
else P ←    /* and convert the others into actions */
    ({(open d) | d ∈ Open & position(d, N) = closed},
     {(close d) | d ∈ Closed & position(d, N) = open})
    add P to the list of admissible level-1 plans
    return
```

Figure 3: Generation of Admissible Level-1 Plans

by the diagnostic reasoner is small enough to be entirely explored. For our network example in Figure 1, it contains most of the time less than a hundred plans. These are all evaluated using a utility function that captures the criteria mentioned in Subsection 2.3, and the best one is returned to the diagnostic reasoner. There might be no admissible level-1 plan that resupplies all non-faulty lines that could be resupplied if we had considered the entire plan space, so we might choose a plan that only constitutes a partial solution. However, this is completely reasonable: other types of plans are rarely used in reality because they require a complex protocol with the dispatching center.

The space of admissible level-1 plans is explored as show in Figure 3. Recall that feeders are trees whose roots are CBs and whose leaves are open SDs. The idea is to make these trees grow up towards the non-faulty unsupplied areas. We examine all possible ways of extending non-faulty bordering feeders towards the cut off feeders, starting at their common leaves, attempting to close them, and stopping the extension by opening a device before reaching a faulty area. Following the same principle, all possible ways of rebuilding part of the cut off feeders starting from their CB-root are examined. Note we cannot always choose to close/open a device to go on/stop the extension: devices whose AC is incorrect and manually controlled devices cannot be operated and must keep their current position. Also, closing a device cannot be chosen if this leads a faulty area to be reached, a line to be fed via multiple CBs, or admissibility to be violated.

E.g., if the feeder fed via CB 1 is cut off, we can choose to extend the feeder fed by CB 5 towards the former, starting by closing their common leaf RSD 53. This commits us to to resupply the line between this latter and MSD 19, its downstream child on the extended feeder. Since MSD 19 is manual and closed, we are forced to go on the extension one step further, so the line between MSD 19 and RSD 18 is also resupplied. Then, we can choose to open RSD 18 to stop the extension or to let it closed, and so on. Other possibilities are to extend the feeder fed by CB 6 starting at RSD 63, and the cut off feeder starting with CB 1. All combination of these extensions are examined.

### 5.3 SAMPLE SESSION

Figure 4 shows a sample session with our prototype. Two faults cause CB 1 to open: one between RSD 11 and the ground (above RSD 11 in Figure 1), and the other between RSDs 16 and 18. Furthermore, the fault detector of RSD 16 and the actuator of RSD 11 lie, and the position detector of RSD 11 is broken. All this is unknown to the prototype, which can solely observe that the FDs of RSDs 12 and 11 are the only ones indicating a fault downstream.

Given this observation, the most probable single-fault location is between RSD 11 and the ground, which implies that the FD of RSD 12 lies. The plan is then simply to open RSD 11 to isolate the fault, and to resupply all the lines by reclosing CB 1. After operating RSD 11, it is unknown whether this one is really open because its PD is broken, and in fact it is still closed because its AC is lying. Thus, when attempting to reclose CB 1, this one opens because it is still feeding the two faults. This leads to a revision of the current plan, which is materialized by a dash line on the figure.

The newly most probable hypothesis is a fault between RSDs 12, 63 and 16. This implies that the FD of RSD 11 is the only liar, which is more probable than a failure of the previous opening operation on RSD 11. The plan is to open RSDs 12 and 16 to isolate the fault, to close RSD 53 to resupply the downstream lines via CB 5, and to close CB 1 to resupply the upstream lines. The plan is executed until the closing operation on RSD 53, at which point CB 5 opens (because it feeds the fault between RSDs 16 and 18).

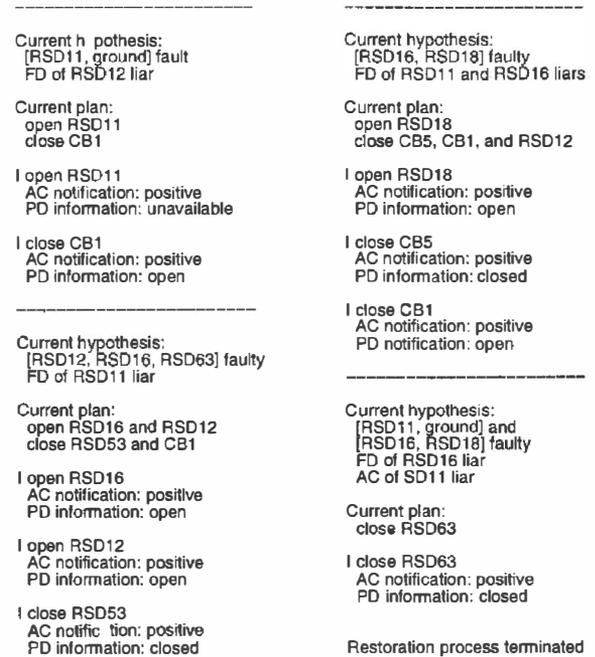

Figure 4: Sample Session

The information returned by all the FDs on the feeder that was fed via CB 5 (all indicate a fault downstream) makes it now sufficiently probable that the fault is between RSD 16 and 18. RSD 16 being already open, it suffices to open RSD 18 to isolate the fault. The rest of the plan is to reclose CBs 5 and 1 and RSD 12 (this latter had been opened at the previous step). When closing CB 1, this one opens because it is feeding the fault above RSD 11.

The past history of actions and observations implies that there must be at least two faults. The diagnostic reasoner generates the two faults hypotheses, the most probable of which is the right one. It implies that the FD of RSD 16 and the AC of RSD 11 are liars. Therefore, it is useless to attempt to open RSD 11 again to isolate the first fault, and we cannot resupply the lines between CB 1 and RSD 12. The second fault is already isolated, since RSDs 12, 16, and 18 have already been opened. Furthermore, the lines downstream of RSD 18 have already been resupplied, so the plan only consists in closing RSD 63 to resupply the lines between RSDs 12, 63, and 16, via CB 6. The execution of this closing operation ends the restoration process.

This session takes less than one minute CPU time on a Sparc 10, and the efficiency of our prototype (implemented in Standard ML) could be greatly improved. A number of other examples involving multiple faults on multiple feeders have been tested and gave satisfactory results (the prototype could reestablish the power within a minute). This suggests that our short term solution can be extended and still cope with real-size networks, which is what we intend to do next. On the one hand, as the sample session makes it obvious, risks are not taken into account at all in the evaluation of a plan, which increases the breakdown costs. On the other hand, all level-1 plans for a given state are generated, and those are never reused when an AC breaks, though it would suffice to look up in the list of those plans to find an appropriate one. Our future prototype will then exploit these generated plans to evaluate the risks with respect to AC failures. At the top-level, risks with respect to a wrong fault localization will be evaluated by projecting plans over a given horizon for some subset of the state hypotheses. Varying the parameters the lookahead will settle a tradeoff between computation time and plan quality, and will indicate how limiting our current prototype was by not handling partial observability at planning time.

## 6   CONCLUSION

Integrating diagnosis and repair and more generally planning for partially observable domains, are two topics that have recently emerged as highly motivating ones in the model-based diagnosis and planning communities. We have presented a real-world problem that confirms the usefulness of making this type of research successful. We believe that many other applications in diagnosis and repair, monitoring of dynamic systems, and planning, share similar properties and are demanding of those technologies.

However, we have shown that the existing approaches are not powerful enough or computationally too expensive for our application, in particular when a general-purpose planner is used as the core or as a subcomponent of the architecture. Our analysis suggests that several factors are responsible for this: (1) planning with domain constraints is expensive but the lack of it restricts the system to too basic types of repair actions (2) acting optimally in partially observable domains is expensive, neglecting partial observability at planning time is not entirely satisfactory, and precise characterizations of a good middle ground according to domains' features are lacking, and (3) as long as subsequent effort is not put in designing high level languages and methods for specifying and exploiting the domains' specificities within general-purpose systems, those are of little use for many real-world applications.

We therefore hope that this paper will motivate further research in both communities. This could be, for instance, in the following respective directions: (1) studying more complex architectures for integrating model-based diagnosis and repair, in particular those that could enable a more equal repartition of the work and of the domain model between the planner and the other components, (2a) extending the algorithms for POMDPs so as to focus on relevant parts of the belief state space, and so as to interleave computation and execution as appropriate (see e.g., [Dean et al., 1993; Tash and Russell, 1994] for the fully observable case), (2b) investigating restricted types of domain constraints and of utility functions that could be efficiently handled by stochastic planners that extend classical planning algorithms, and (3) looking more carefully for theoretical foundations that could enable us to exploit domains' specificities in planning, such as those of decision theory.

## References

[Beschta et al., 1993] A. Beschta, O. Dressler, H. Freitag, M. Montag, and P. Struß. A model-based approach to fault localisation in power transmission networks. *Intel. Syst. Eng.*, 1-2:190–201, 1993.

[Bredillet et al., 1994] P. Bredillet, I. Delouis, P. Eyrolles, O. Jehl, J.-P. Krivine, and P. Thiault. The AUSTRAL expert system for power restoration on distribution systems. In *Proc. Int. Conf. on Intelligent Systems Application to Power Systems*, pages 295–302. EC2, 1994.

[Cassandra et al., 1994] A. Cassandra, L. Kaelbling, and M. Littman. Acting optimally in partially observable stochastic domains. In *Proc. AAAI-94*, pages 1023–1028, 1994.

[Dean et al., 1993] T. Dean, L. P. Kaelbling, J. Kirman, and A. Nicholson. Planning with deadlines in stochastic domains. In *Proc. AAAI-93*, pages 574–579, 1993.

[Draper et al., 1994] D. Draper, S. Hanks, and D. Weld. A probabilistic model of action for least-commitment planning with information gathering. In *Proc UAI-94*, pages 178–186, 1994.

[Friedrich and Nejdl, 1992] G. Friedrich and W. Nejdl. Choosing observations and actions in model-based diagnosis-repair systems. In *Proc. DX-92*, pages 76–85, 1992.

[Kushmeric et al., 1994] N. Kushmeric, S. Hanks, and D. Weld. An algorithm for probabilistic least-commitment planning. In *Proc. AAAI-94*, 1994.

[Mondon et al., 1991] E. Mondon, B. Heilbronn, Y. Harmand, O. Paillet, and H. Fargier. MARS: an aid network restoration after a local disturbance. In *IEEE Conference on Power Industry Computer Applications*, 1991.

[Sun and Weld, 1992] Y. Sun and D. Weld. Beyond simple observation: Planning to diagnose. In *Proc. DX-92*, pages 67–75, 1992.

[Tash and Russell, 1994] J. Tash and S. Russell. Control strategies for a stochastic planner. In *Proc. AAAI-94*, pages 1079–1085, 1994.

[Thiébaux et al., 1993] S. Thiébaux, J. Hertzberg, W. Shoaff, and M. Schneider. A stochastic model of actions and plans for anytime planning under uncertainty. In *Proc. EWSP-93*, pages 292–305, 1993.

[Thiébaux, 1995] S. Thiébaux. *Contribution à la planification sous incertitude et en temps contraint*. PhD thesis, Université de Rennes I, June 1995. In French.